%% file: paper.tex
\documentclass{article}
\usepackage{spconf,amsmath,graphicx}
\usepackage{booktabs}

\def\x{{\mathbf x}}
\def\y{{\mathbf y}}
\def\h{{\mathbf h}}

\DeclareMathOperator*{\argmax}{argmax}
\def\ci{\perp\!\!\!\perp}

\title{Exploring Architectures, Data and Units For Streaming End-to-End Speech Recognition with RNN-Transducer}

\name{Kanishka Rao, Ha\c{s}im Sak, Rohit Prabhavalkar}
\address{Google Inc.,\\
	Mountain View, CA, U.S.A.\\
	\texttt{\{kanishkarao,hasim,prabhavalkar\}@google.com}}

\begin{document}
\maketitle

\input{abstract}
\input{intro}

\input{rnnt}

\input{training}
\input{experiments}
\input{results}
\input{analysis}
\input{conclusion}
\input{acknowledgements}

\bibliographystyle{IEEEbib}
\bibliography{paper}

\end{document}

%% file: abstract.tex
\begin{abstract}

We investigate training end-to-end speech recognition models with the recurrent neural network
transducer (RNN-T): a streaming, all-neural, sequence-to-sequence architecture which jointly
learns acoustic and language model components from transcribed acoustic data.
We explore various model architectures and demonstrate how the model can be improved further if additional text or
pronunciation data are available. The model consists of an `encoder', which is initialized
from a connectionist temporal classification-based (CTC) acoustic model, and a
`decoder' which is partially initialized from a recurrent neural network language model trained on text data alone.
The entire neural network is trained with the RNN-T loss and directly outputs the recognized transcript
as a sequence of graphemes, thus performing end-to-end speech recognition. We find that performance
can be improved further through the use of sub-word units (`wordpieces') which capture longer context
and significantly reduce substitution errors. The best RNN-T system, a twelve-layer LSTM encoder with a
two-layer LSTM decoder trained with 30,000 wordpieces as output targets achieves a word error rate of
8.5\% on voice-search and 5.2\% on voice-dictation tasks and is comparable to a
state-of-the-art baseline at 8.3\% on voice-search and 5.4\% voice-dictation.

\end{abstract}


%% file: intro.tex
\section{Introduction \label{sec:introduction}}
The current state-of-the-art automatic speech recognition (ASR) systems break down
the ASR problem into three main sub-problems: acoustic, pronunciation and language modeling.
Speech recognition involves determining the most likely word sequence,
$W=w_1,...,w_n$, given an acoustic input sequence, $\mathbf{x}=x_1,...,x_T$,
where $T$ represents the number of frames in the utterance:
\begin{equation}
	W^{*} = \argmax_W P(W|\x),
\end{equation}
which is typically decomposed into three separate models, as follows:
\begin{align}
	W^{*} &= \argmax_W \sum_\phi P(\x, \phi | W) P(W) \\
	      &\approx \argmax_{W, \phi} p(\x|\phi) P(\phi|W) P(W)
\end{align}
The acoustic model, $p(\x|\phi)$, predicts the likelihood of the acoustic input
speech utterance given a phoneme sequence, $\phi$; for conditional models that
directly predict $P(\phi|\x)$, the likelihood is typically replaced with a
scaled likelihood obtained by dividing the posterior with the prior, $P(\phi)$,
in so-called hybrid models~\cite{MorganBourlard95}.
Deep recurrent neural networks with long short-term memory (LSTM)
cells~\cite{lstm} have recently been shown to be ideal for this
task~\cite{lstm_am1, lstm_am2, lstm_am3}.
The pronunciation model, $P(\phi|W)$, is typically built from pronunciation
dictionaries curated by expert human linguists, with back-off to a
grapheme-to-phoneme (G2P) model~\cite{g2p} for out of dictionary words.
Finally, an N-gram model trained on text data may be used as a language model,
$P(W)$.

Recently, there has been considerable interest in training end-to-end models for
ASR~\cite{las, baidu, BahdanauChorowskiSerdyukEtAl16}, which directly output
word transcripts given the input audio.\footnote{ In the context of this work,
we consider models that are all-neural, and directly output word transcripts from
audio utterances as being end-to-end.}
Thus, these models are much simpler than conventional ASR systems as a single
neural network can be used to directly recognize utterances, without requiring
separately-trained acoustic, pronunciation and language model components.
A particular class of architecures known as sequence-to-sequence
models~\cite{seq2seq} are particularly suited for end-to-end ASR as they include
an \emph{encoder} network which corresponds to the acoustic model of a
conventional system and a decoder network which corresponds to the language
model.

One drawback of typical encoder-decoder type architectures (e.g.,~\cite{las,
BahdanauChorowskiSerdyukEtAl16}) is that the entire input sequence is encoded
before the output sequence may be decoded and thus these models cannot be used
for real-time streaming speech recognition.
Several streaming encoder-decoder architectures have been proposed previously,
including the neural transducer~\cite{nt}, the recurrent neural aligner
(RNA)~\cite{rna}, and the recurrent neural network transducer
(RNN-T)~\cite{rnnt1, rnnt2}.
In particular, these architectures allow the output to be decoded as soon as the first
input is encoded, without introducing additional latency incurred when
processing the entire utterance at once.
In this work we only consider streaming recognition architectures, specifically
the RNN-T model.

Despite recent work on end-to-end ASR, conventional systems still remain the
state-of-the-art in terms of word error rate (WER) performance.
For example, in our previous work~\cite{rohit1} we evaluated a number of
end-to-end models including attention-based models~\cite{las} and
RNN-T~\cite{rnnt1, rnnt2} trained on $\sim$12,500 hours of transcribed training
data; although end-to-end approaches were found to be comparable to a
state-of-the-art context-dependent phone-based baseline on dictation test sets,
these models were found to be significantly worse than the baseline on
voice-search test sets.
End-to-end systems are typically trained using transcribed acoustic data sets,
which are relatively expensive to generate and thus much smaller than text-only
data sets, which are used to train LMs in a traditional speech recognizer.
A deficiency of end-to-end systems appears to be in their language modeling
capacity~\cite{rohit1} which may be because large text-only data are not
utilized in end-to-end systems.

In this work we explore a particular sequence-to-sequence architecure, RNN-T,
and show how text and pronunciation data may be included to improve end-to-end
ASR performance.
Another contribution of this work is to investigate the use of
wordpieces~\cite{wp}, which have been explored previously in the context of
machine translation, as a sub-word unit for end-to-end speech recognition.

The paper is organized as follows: in Section~\ref{sec:rnnt} we describe the RNN-T and
how it may be used for streaming recognition. Section~\ref{sec:training} describes how the RNN-T
is trained including the units, architectures and pre-training parts of the model. The experimental
setup including the baseline system are detailed in Section~\ref{sec:experiments}. Section~\ref{sec:results}
compares the word error rate performance of various RNN-T models and the baseline to show relative improvement.
We find that the techniques introduced in this work mostly improve the language modeling of the RNN-T,
Section~\ref{sec:analysis} shows some select examples of such improved recognition. A concluding summary and
acknowledgements are in Section~\ref{sec:conclusion} and Section~\ref{sec:acknowledgements}.

%% file: rnnt.tex
\section{RNN-Transducer}
\label{sec:rnnt}
\begin{figure}
	\centering
	\includegraphics[width=0.5\columnwidth]{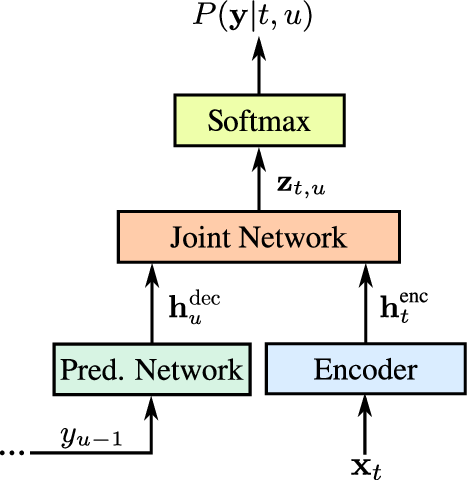}
	\caption{The RNN-T model. The model consists of an encoder network,
	which maps input acoustic frames into a higher-level representation, and
	a prediction and joint network which together correspond to the decoder
	network. The decoder is conditioned on the history of previous
	predictions.}
	\label{fig:rnnt}
\end{figure}

The RNN-T was proposed by Graves~\cite{rnnt1} as an extension to the
connectionist temporal classification (CTC)~\cite{ctc} approach for sequence
labeling tasks where the alignment between the input sequence, $\x$, and the
output targets $\y$ is unknown.
This is accomplished in the CTC formulation by introducing a special label,
called the \emph{blank} label, which models the probability of outputting no
label corresponding to a given input frame.
CTC has been widely used in previous works to train end-to-end ASR
models~\cite{baidu, e2e_ctc1, e2e_ctc2}.
However, a major limitation of CTC is its assumption that model outputs at a
given frame are independent of previous output labels: $y_t \ci y_j | \x$, for
$t < j$.

The RNN-T model, depicted in Figure~\ref{fig:rnnt}, consists of an
\emph{encoder} (referred to as the transcription network in~\cite{rnnt1}), a
prediction network and a joint network; as described in~\cite{rohit1}, the RNN-T
model can be compared to other encoder-decoder architectures such as ``listen,
attend, and spell"~\cite{las}, if we view the combination of the prediction
network and the joint network as a decoder.
The encoder is an RNN which converts the input acoustic frame $\x_t$ into a
higher-level representation, $\h^\text{enc}_t$, and is analogous to a CTC-based
AM in a standard speech recognizer.
Thus, as in CTC, the output of the encoder network, $\h^\text{enc}_t$, is
conditioned on the sequence of previous acoustic frames $x_0, \cdots, x_t$.
\begin{equation}
\h_{t}^\text{enc} = f^\text{enc}(x_t),
\end{equation}

The RNN-T removes the conditional independence assumption in CTC by introducing
a \emph{prediction network}, an RNN that is explicitly conditioned on the
history of previous non-blank targets predicted by the model.
Specifically, the prediction network receives as input the last \emph{non-blank}
label, $y_{u-1}$, to produce as output $\h^\text{dec}_u$.
\begin{equation}
\h_{u}^\text{dec} = f^\text{dec}(y_{u-1}).
\end{equation}

Finally, the \emph{joint network}, is a feed-forward network that combines the
outputs of the prediction network and the encoder to produce logits
($\mathbf{z}_{t, u}$) followed by a softmax layer to produce a distribution
over the next output symbol (either the blank symbol or one of the
output targets).
\begin{equation}
z_{t,u} = f^\text{joint}(\h_{t}^\text{enc}, \h_{u}^\text{dec})
\end{equation}
We use the same form for $f^\text{joint}$ as described in~\cite{rnnt2}.
The entire network is trained jointly to optimize the RNN-T loss~\cite{rnnt1},
which marginalizes over all alignments of target labels with blanks as in CTC,
and is computed using dynamic programming.

During each step of inference, the RNN-T model is fed the next acoustic frame
$\x_t$ and the previously predicted label $y_{u-1}$, from which the model produces
the next output label probabilities $P(y|t, u)$. If the predicted label, $y_{u}$, is non-blank,
then the prediction network is updated with that label as input to generate 
the next output label probabilities $P(y|t, u + 1)$. Conversely, if
a blank label is predicted then the next acoustic frame, $\x_{t+1}$, is used to update the encoder
while retain the same prediction network output resulting in $P(y|t + 1, u)$. In this way
the RNN-T can stream recognition results by alternating between updating the encoder and the prediction
network based on if the predicted label is a blank or non-blank. 
Inference is terminated when blank is output at the last frame, $T$.

During inference, the most likely label sequence is computed using beam search
as described in~\cite{rnnt1}, with a minor alteration which was found to make
the algorithm less computationally intensive without degrading performance: we
skip summation over prefixes in \texttt{pref$(\mathbf{y})$} (see Algorithm
1 in~\cite{rnnt1}), unless multiple hypotheses are identical.

Note that unlike other streaming encoder-decoder architectures such as
RNA~\cite{rna} and NT~\cite{nt}, the prediction network is not conditioned on
the encoder output.
This allows for the the pre-training of the decoder as a RNN language model on
text-only data as described in Section~\ref{sec:training}.

%% file: training.tex
\section{Units, Architectures and Training}
\label{sec:training}
\begin{figure*}[!t]
\begin{center}
\includegraphics[width=\textwidth]{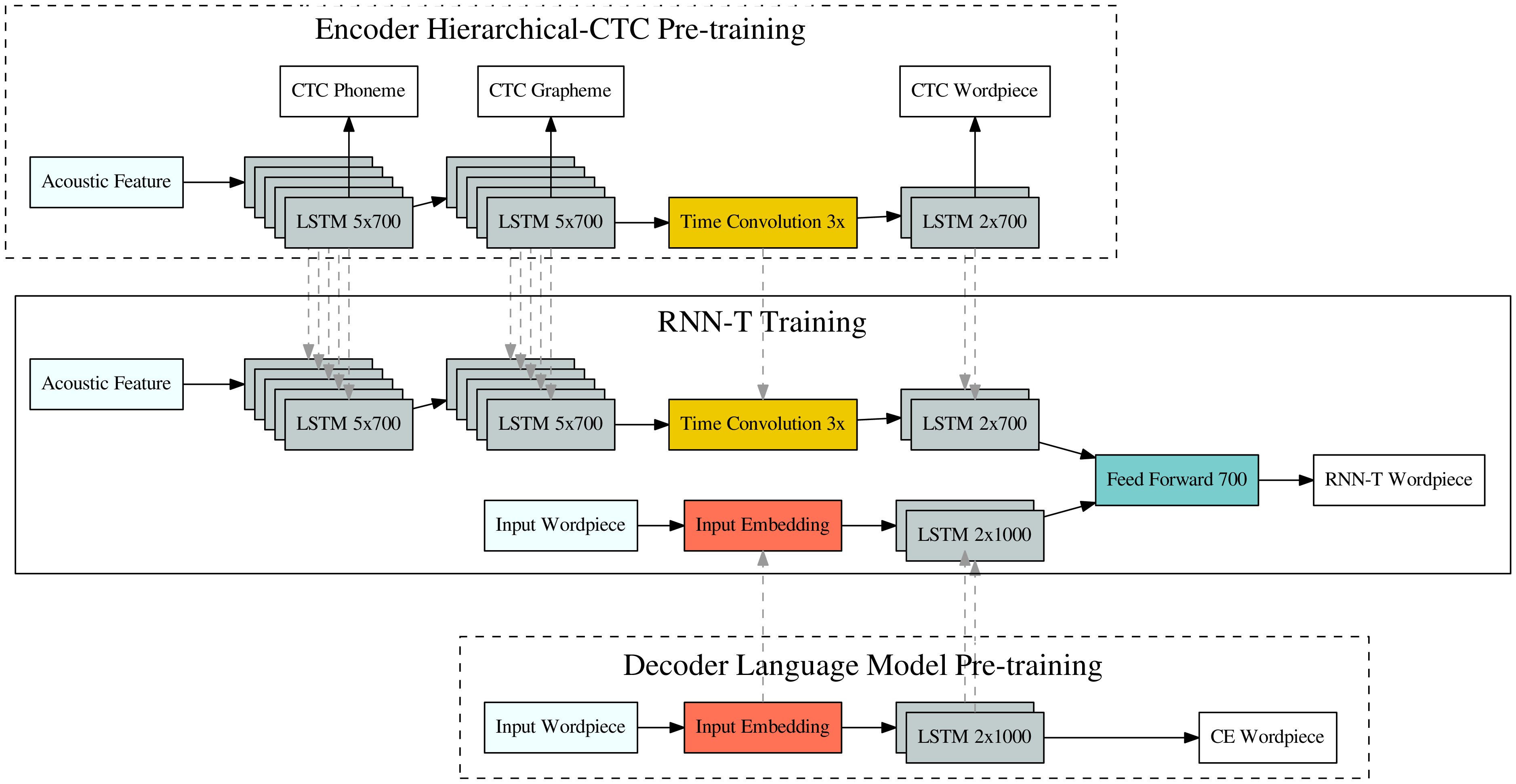}
\end{center}
\caption{The various stages of training a wordpiece RNN-T. The encoder network is pre-trained as a hierarchical-CTC network simultaneously
predicting phonemes, graphemes and wordpieces at 5, 10 and 12 LSTM layers respectively. A time convolutional layer reduces the encoder
time sequence length by a factor of three. The decoder network is trained as a LSTM langauge model predicting wordpieces optimized with
a cross-entropy loss. Finally, the RNN-T network weights are initialized from the two pre-trained models, indicated by the dashed lines,
and the entire network is optimized using the RNN-T loss.}
\label{rnnt}
\end{figure*}

We investigate the use of graphemes and sub-words (wordpieces) as output lexical units in RNN-T models.
For the graphemes, we use letters (\texttt{a-z}), digits (\texttt{0-9}), special
symbols (\texttt{\&.'\%/-:}) and a space symbol (\texttt{$<$space$>$}).
The space symbol is used for segmenting recognized grapheme sequences to word sequences.

State-of-the-art large vocabulary speech recognition systems recognize millions of different words,
inference for RNN-T with that many output labels would be impractically slow.
Therefore, as subword units, we use wordpieces as described in ~\cite{wp}.
We train a statistical wordpiece model with word counts obtained from text data for segmenting each word individually into subwords.
An additional space symbol is included in subword units.
An example segmentation for the sentence \texttt{tortoise and the hare} is
\texttt{$<$tor$>$ $<$to$>$ $<$ise$>$ $<$space$>$ $<$and$>$ $<$space$>$ $<$the$>$ $<$space$>$ $<$ha$>$ $<$re$>$}.
Wordpieces have be shown to benefit end-to-end recognition~\cite{lsd} since they offer
a balance with longer context than graphemes and a tunable number of labels.
Since the wordpiece
model is based on word frequencies, more common words appear as a single label.
A vocabulary of 1,000 generated wordpieces
includes words like `mall', `remember' and `doctor' while a vocabulary of 30,000 wordpieces also includes
less common words like `multimedia', `tungsten' and `49er'. The wordpiece models may also output
any word that the grapheme model may; we find that all the graphemes are included in the wordpiece vocabularies.

For the encoder networks in RNN-T models, we experimented with deep LSTM networks (5 to 12 layers).
For the decoder networks, we used a stack of 2 layer LSTM network, a feed-forward layer and a softmax layer.
In addition to training models with random initialization of parameters, we explored variations of initializing encoder and decoder network parameters from pre-trained models.
It has been previously shown that initializing RNN-T encoder parameters from a model trained with the CTC loss is beneficial for the phoneme recognition task~\cite{rnnt2}.
We experimented with initializing encoder networks from models trained with the
CTC loss and with initializing LSTM layer parameters in prediction networks from LSTM language models trained on text data.
After initialization of encoder and prediction network weights from separate pre-trained models, the entire RNN-T model weights are trained with the RNN-T objective.

We show one example architecture for the RNN-T wordpiece model in Figure 2.
The figure also shows the pre-trained CTC LSTM acoustic model and LSTM language model architectures used to initialize the encoder and prediction network weights.
The dotted arrows indicate the pre-trained layers used to initialize specific layers in the RNN-T model.
The encoder networks in RNN-T models are pre-trained with the CTC loss using phonemes, graphemes and wordpieces as output units.
We investigate encoder architectures with multi-task training using hierarchical-CTC~\cite{hctc} with various
'hierarchies' of CTC losses at various depths in the encoder network. With hierarchical-CTC
the encoder networks are trained with multiple simultaneous CTC losses which was
beneficial for grapheme recognition~\cite{hctc_g}. After pre-training all CTC losses
and additional weights associated with generating softmax probabilities are discarded.
For the wordpiece models which have longer duration than graphemes, we employ an additional 'time-convolution' in the encoder network to reduce the
sequence length of encoded activations which is similar to the pyramidal sequence length reduction in ~\cite{las}.
For these models, we used filters covering 3 non-overlapping consecutive activation vectors, thus reducing them to a single activation vector.
The LSTM layers in decoder networks are pre-trained as a language model using the graphemes or wordpieces as lexical units.
The input to the network is a label (grapheme or wordpiece) in a segmented sentence represented as a one-hot vector.
The target for the network is the next label in the sequence and the model is trained with the cross-entropy loss.
The weights in the softmax output layer are discarded after pre-training and only the LSTM network weights are used to partially initialize the RNN-T prediction network.
For wordpiece language models, we embed labels to a smaller dimension.
These embedding weights are also used to initialize the RNN-T wordpiece models.

%% file: experiments.tex
\section{Experimental Setup} \label{sec:experiments}

We compare the RNN-T end-to-end recognizer with a conventional ASR system consisting of separate acoustic, pronunciation and language models.
The acoustic model is a CTC trained LSTM that predicts context-dependent (CD) phonemes
first fine-tuned with sequence discriminative training as described in~\cite{lstm_am3} and
further improved with word-level edit-based minimum Bayes risk (EMBR) proposed recently by Shannon~\cite{embr}.
Acoustic models are trained on a set of $\sim$22 million hand-transcribed anonymized
utterances extracted from Google US English voice traffic, which corresponds to $\sim$18,000
hours of training data. These include voice-search as well as voice-dictation utterances.
We use 80-dimensional log mel filterbank energy features computed every 10ms stacked every 30ms to a single 240-dimensional acoustic
feature vector. To achieve noise robustness acoustic training data is distorted as
described in~\cite{lstm_am}. The pronunciation model is a dictionary containing hundreds of thousands of human expert transcribed US English word pronunciations.
Additional word pronunciations are learned from audio data using pronunciation 
learning techniques~\cite{pronlearning}. For out-of-dictionary words a G2P model is trained using transcribed word pronunciations.
A 5-gram language model is trained  with a text sentence dataset which includes untranscribed anonymized speech logs:
150 million sentences each from voice-search and voice-dictation queries, and anonymized typed logs including tens of billion
sentences from Google search from various sources. The language model is pruned to 100-million n-grams with a target vocabulary of 4 million
and the various sources of text data are re-weighted using interpolation~\cite{interpolation} for the
optimal word error rate performance. Single-pass decoding with a conventional WFST is carried out to generate recognition transcripts.

The RNN-T is trained with the same data as the baseline. The CTC encoder network is pre-trained with
acoustic transcribed data and as with the baseline acoustic model the pronunciation model is used to
generate phoneme transcriptions for the acoustic data. The RNN-T decoder is pre-trained on the text only data as a LSTM language model, 
roughly half a billion sentences from the text data are sampled according to their count and the data source 
interpolation weight (as optimized in the baseline). All RNN-T models are trained with LSTM networks in the tensorflow~\cite{tensorflow} toolkit with
asynchronous stochastic gradient descent.  Models are evaluated using the RNN-T beam search algorithm
with a beam of 100 for grapheme models and 25 for wordpiece models and a temperature of 1.5 on the softmax.
Word error rate (WER) is reported on a voice-search and a voice-dictation test set with roughly 15,000 utterances each.

%% file: results.tex
\begin{table*}[!t]
  \caption{Word error performance on the voice-search and dictation tasks for various RNN-T trained with
  graphemes and wordpieces with various architectures and pre-training. Also shown for each model is which
  types of training data are included: acoustic, pronunciation or text. The baseline is a state-of-the-art
  conventional speech recognition system with separate acoustic, pronunciation and language models trained
  on all available data. The parameters for the baseline system include 20 million weights from the acoustic model
  network, 0.2 million for each word in the pronunciation dictionary and the 100 million n-grams in the language model.}
  \label{wer}
  \centering
  \begin{tabular}{lcccccccccc} \toprule
          & \multicolumn{2}{c}{Layers} &\multicolumn{2}{c}{Pre-trained}   &   \multicolumn{3}{c}{Training Data Used}& & \multicolumn{2}{c}{WER(\%)} \\
    Units & Encoder & Decoder & Encoder & Decoder & Acoustic & Pronunciation & Text & Params & VS & IME \\ \hline \midrule
    \multicolumn{11}{c}{\textbf{RNN-T}}  \\
    Graphemes & 5x700 & 2x700 &no & no & yes & no & no & 21M & 13.9 & 8.4 \\
    Graphemes & 5x700 & 2x700 &yes & no & yes & no & no & 21M & 13.2 & 8.0 \\
    Graphemes & 8x700 & 2x700 &yes & no & yes & no & no & 33M & 12.0 & 6.9 \\
    Graphemes & 8x700 & 2x700 &yes & no& yes & yes & no & 33M & 11.4 & 6.8 \\
    Graphemes & 8x700 & 2x700 &yes & yes& yes & yes & yes & 33M & 10.8 & 6.4 \\  \hline
    Wordpieces-1k & 12x700 & 2x700&yes & yes & yes & yes & yes & 55M & 9.9 & 6.0 \\
    Wordpieces-10k & 12x700 & 2x700&yes & yes & yes & yes & yes & 66M & 9.1 & 5.3 \\
    Wordpieces-30k & 12x700 & 2x1000&yes & yes & yes & yes & yes & 96M & 8.5 & 5.2 \\ \hline  \hline
    \multicolumn{11}{c}{\textbf{Baseline}}  \\
    - & - & - & -& - & yes & yes & yes & 120.2M & 8.3 & 5.4 \\ \hline
 \bottomrule
  \end{tabular}
\end{table*}

\section{Results}
\label{sec:results}

We train and evaluate various RNN-T and incrementally show the WER impact with
each improvement.

A grapheme based RNN-T is trained from scratch (no pre-training) on the acoustic data
with a 5-layer LSTM encoder of 700 cells and a 2-layer LSTM decoder of 700 cells. A final 700 unit
feed-forward layer and a softmax layer output grapheme label probabilities. We compare this model
to a model with identical architecture but with the encoder CTC pre-trained. We find CTC pre-training
to be helpful improving WER 13.9\%$\rightarrow$13.2\% for voice-search and 8.4\%$\rightarrow$8.0\%
for voice-dictation.

A model with a deeper 8-layer encoder is also trained with a multi-CTC loss at depth 5 and depth 8 where both
losses are optimized for the same grapheme targets. We found training 8-layer models without a
multi-loss setup to be unstable which we acknowledge may be addressed with recent advancements in
training deeper recurrent models~\cite{highway} but are not tested as part of this work.
The deeper 8-layer encoder further improves WER 13.2\%$\rightarrow$12.0\% for voice-search and 8.4\%$\rightarrow$6.9\%
for voice-dictation.

To incorporate the knowledge of phonemes and specifically the pronunciation dictionary data we train a 8-layer
encoder with hierarchical-CTC with a phoneme target CTC at depth 5 and a grapheme target CTC at depth 8.
In this way the network is forced to model phonemes and is exposed to pronunciation variants in the labels where
the same word (and thus same grapheme sequence) may have different pronunciations (and thus phoneme sequences).
This approach does not address including pronunciations for words that do not occur in the acoustic training
data, which we leave as future work. We find that the pronunciation data improves WER 12.0\%$\rightarrow$11.4\% for voice-search
but with little improvement for voice-dictation. Unlike voice-search the voice-dictation test set is 
comprised of mostly common words, we conjecture that it may be sufficient to learn pronunciations for these
words from the acoustic data alone and thus may not benefit from additional human transcribed pronunciations.

Next, to include the text data we pre-train a 2-layer LSTM with 700 cells as a language model with grapheme targets. The model
is trained until word perplexity on a held-out set no longer improves, Table~\ref{lmperp} shows the word preplexity and
sizes of the various language models that were trained. Addition of text data in this way improves WER 
11.4\%$\rightarrow$10.8\% for voice-search and 6.8\%$\rightarrow$6.4\% for voice-dictation.

We explore modeling wordpieces, with 1k, 10k and 30k wordpieces, instead of graphemes and make several changes to the architecture. 
The wordpiece encoder network is a 12-layer LSTM with 700 cells each, trained with hierarchical-CTC with phoneme targets at depth 5, graphemes at depth 10
and wordpieces at depth 12. Since wordpieces are longer units we include a time convolution after depth 10 reducing the 
sequence length by a factor of 3. We find that this time convolution does not affect WER but drastically reduces training and
inference time as there are 3 times fewer encoder features that need to be processed by the decoder network. Wordpiece language models
are trained similar to graphemes, since the numbers of labels are much larger an additional input embedding of size 500 is used for wordpiece
models. The wordpiece language models perform much better in terms of word perplexity (Table~\ref{lmperp}) and the RNN-T initialized from
them also see significant WER improvements  (Table~\ref{wer}). The best end-to-end RNN-T with 30k wordpieces achieves a 
WER of 8.5\% for voice-search and 5.2\% on voice-dictation which is on par with the state-of-the-art baseline speech recognition system.

\begin{table}[!h]
  \caption{The number of parameters (in millions) and word perplexity for LSTM language model trained with different units evaluated
  on a held-out set.}
  \label{lmperp}
  \centering
  \begin{tabular}{lcc} \toprule
    Units & Params & Perplexity \\ \hline \midrule
    Graphemes & 6M & 185 \\
    Wordpieces-1k & 10M & 138 \\
    Wordpieces-10k & 20M & 130 \\
    Wordpieces-30k & 59M & 119 \\
 \bottomrule
  \end{tabular}
\end{table}

%% file: analysis.tex
\section{Analysis}
\label{sec:analysis}

We observe that a large part of the improvements described in this work are from a reduction in substitution errors.
Using wordpieces instead of graphemes results in an absolute 2.3\% word error rate improvement,
of this 1.5\% is due to fixing substitution errors. Inclusion of pronunciation and text data
improve voice-search word error rate by an absolute 0.6\% and 0.6\% respectively, all of these are
due to improvements in word substitution errors. Many of the corrected substitution errors seem to be from improved language modeling: words which
may sound similar but have different meaning given the text context. Some selected examples include
improvements with proper nouns: `barbara stanwick' recognized by a grapheme model is fixed when using
wordpieces to the correct name `barbara stanwyck'. Similar improvements are found when including
pronunciation data: `sequoia casino' to `sycuan casino', `where is there' to `where is xur'
and also when including text data: `soldier boy' to `soulja boy', `lorenzo llamas' to `lorenzo lamas'.
We also find that wordpieces capture longer range language context than graphemes in improvements like
`tortoise and the hair' to `tortoise and the hare'.

%% file: conclusion.tex
\section{Conclusion}
\label{sec:conclusion}

We train end-to-end speech recognition models using the RNN-T which predicts graphemes
or wordpieces and thus directly outputs the transcript from audio. We find pre-training the RNN-T
encoder with CTC results in a 5\% relative WER improvement, and using a deeper
8-layer encoder instead of a 5-layer encoder further improves WER by 10\% relative.
We incorporate pronunciation data using a pre-training hierarchical-CTC loss which
includes phoneme targets and find this improves the voice-search WER by 5\% relative
with little impact on the voice-dictation task. To include
text-only data we pre-train the recurrent network in the decoder as LSTM language models resulting
in a overall 5\% relative improvement. We train wordpiece RNN-Ts with 1k, 10k and 30k wordpieces targets
and find that they significantly outperform the grapheme-based RNN-Ts. For comparison we use a baseline
speech recognizer with individual acoustic, pronunciation and language models with state-of-the-art WERs
of 8.3\% on voice-search and 5.4\% on voice-dictation. With a 30k wordpiece RNN-T achieving
WERs of 8.5\% on voice-search and 5.2\% on voice-dictation we demonstrate that a single end-to-end neural model
is capable of state-of-the-art streaming speech recognition.

%% file: acknowledgements.tex
\section{Acknowledgements}
\label{sec:acknowledgements}

The authors would like to thank our colleagues: Fran\c{c}oise Beaufays,
Alex Graves and Leif Johnson for helpful research discussions and Mike Schuster for help with wordpiece models.